\documentclass{article}
\PassOptionsToPackage{numbers,sort&compress}{natbib}
\usepackage[final,sglblindworkshop]{neurips_2026}
\workshoptitle{Continual Learning in the Era of Foundation Models and Embodied Agents}
\usepackage[utf8]{inputenc}
\usepackage[T1]{fontenc}
\usepackage{amsmath,amssymb,amsfonts,booktabs,graphicx,microtype,float}
\usepackage{algorithm,algpseudocode}
\usepackage{tikz}
\usetikzlibrary{arrows.meta,positioning,shapes.geometric}
\usepackage{hyperref,url}
\hypersetup{hidelinks,pdfauthor={Zhentao Xu},pdftitle={PID-Gated Continual Prompt Adaptation Under Noisy Feedback}}
\title{PID-Gated Continual Prompt Adaptation Under Noisy Feedback}
\author{%
  \href{https://orcid.org/0009-0000-5827-7879}{Zhentao Xu}\\
  Independent Researcher\\
  Sunnyvale, CA, USA\\
  \texttt{frankxu@umich.edu}%
}
\begin{document}
\maketitle

\begin{abstract}
Deploying LLM agents in non-stationary environments requires continually revising their instructions as the task drifts, yet rewriting after every reported mistake can turn noisy feedback into persistent errors. This raises an often-overlooked scheduling question in prompt adaptation: \emph{when} should feedback trigger a lasting instruction update, and at what editor-call cost? Taking inspiration from proportional-integral-derivative (PID) control, we introduce a family of lightweight gates that react to the current error, accumulate evidence over time, and roll back deterioration after an edit. We benchmark these gates against both simple baselines and advanced prompt optimizers (TextGrad and GEPA), over a diverse suite of drifting tasks, LLM parameter sizes, feedback-noise levels, and random seeds. PID-gated continual prompt adaptation yields a more favorable Pareto trade-off between task performance (high accuracy, low error) and editor-call budget than the baselines considered, pointing to control-theoretic scheduling as a simple, general tool for robust prompt adaptation under unreliable feedback. We open source our code at \url{https://github.com/zhentao-xu/pid-gated-agent-learning}.
\end{abstract}

\section{Introduction}
Consider an agent guessing a hidden integer under noisy high/low feedback: it holds a belief about the plausible range $[L,U]$ that must tighten as evidence accumulates, yet a single mislabeled ``higher'' reply can widen the wrong bound and lock in errors for many rounds. This is the general dilemma of adapting instructions in a changing environment: revising after every reported mistake can turn a useful rule into a persistent source of errors, while waiting leaves the agent following an outdated rule. The challenge is deciding when a mistake warrants a lasting change.

Language models offer several ways to learn from feedback: revising an answer \cite{madaan2023selfrefine}, retaining verbal experience across attempts \cite{shinn2023reflexion}, or optimizing prompts through textual feedback \cite{yuksekgonul2025textgrad} and candidate search \cite{yang2024opro,agrawal2025gepa}. In each, however, the scheduling decision is fixed: revisions fire after every attempt or after every observed error subject to a call ceiling, offering no principled way to withhold an edit when the feedback itself might be wrong. We instead ask a complementary scheduling question: \emph{when should feedback trigger a persistent instruction update?} This connects prompt adaptation to concept drift \cite{gama2014survey} and continual learning \cite{wu2024continual}.

Proportional-integral-derivative (PID) control \cite{astrom1995pid} suggests three complementary responses: reacting to current error, accumulating evidence over time, and monitoring deterioration after an edit \cite{astrom2008feedback}. We translate this intuition into discrete editing gates and compare them against online TextGrad and GEPA using one LLM-based editor and structured predictors. Across our empirical study (Section~\ref{sec:results}, Table~\ref{tab:benchmark-main}, Figure~\ref{fig:performance-vs-calls}), the PID gates collectively yield a more favorable Pareto trade-off between task performance and editor-call budget than the baselines.

\section{Related Work}
\label{sec:related}

\paragraph{Self-evolving LLM agents and noisy evaluators.} Methods that improve LLM behavior from natural-language feedback split by \emph{what} they update. In the textual loop, Self-Refine \cite{madaan2023selfrefine}, Reflexion \cite{shinn2023reflexion}, TextGrad \cite{yuksekgonul2025textgrad}, OPRO \cite{yang2024opro}, and GEPA \cite{agrawal2025gepa} revise prompts, chains-of-thought, or candidate instructions in response to observed errors, and skill-library agents such as Voyager \cite{wang2024voyager} and SkillWeaver \cite{zheng2025skillweaver} accumulate reusable textual skills. In the parameter loop, STaR \cite{zelikman2022star}, Self-Rewarding Language Models \cite{yuan2024self}, RLHF \cite{ouyang2022training}, and Constitutional AI \cite{bai2022constitutional} convert feedback into weight updates. Our PID gates target the textual loop, where every edit is a stochastic LLM rewrite of a persistent artifact, so the scheduling question of \emph{when} to invoke the editor is especially acute. The feedback signal itself is unreliable in both regimes: LLM-as-judge suffers position, verbosity, and self-preference biases \cite{zheng2023judging,wang2024fair,panickssery2024llm}, sycophancy \cite{sharma2024sycophancy}, and models cannot reliably self-correct \cite{huang2024selfcorrect}. This motivates evidence-gated editing rather than reacting to every reported error.

\paragraph{Control theory for learning and continual adaptation.} Feedback control has long informed machine learning: SGD-momentum as PI control \cite{an2018pid}, KL-weight scheduling in VAEs as PID regulation \cite{shao2020controlvae}, constrained RL as a PID-Lagrangian \cite{stooke2020responsive}, and deep-learning optimizers as continuous-time PID \cite{chen2024accelerated}. Our work instantiates PID intuition \emph{outside} the parameter space of a differentiable model, on natural-language artifacts driven by a stochastic LLM editor. On the continual-learning side, catastrophic forgetting \cite{mccloskey1989catastrophic} and the stability-plasticity dilemma have driven decades of research \cite{parisi2019continual,wang2024comprehensive,wu2024continual,gama2014survey}. Classical parameter-space defenses (elastic weight consolidation \cite{kirkpatrick2017overcoming}, replay memory \cite{lopezpaz2017gradient}) have natural prompt-space analogues in the buffered evidence and rollback rules used here.

\section{Method: PID-Gated Adaptation}
\label{sec:method}
\subsection{System overview}
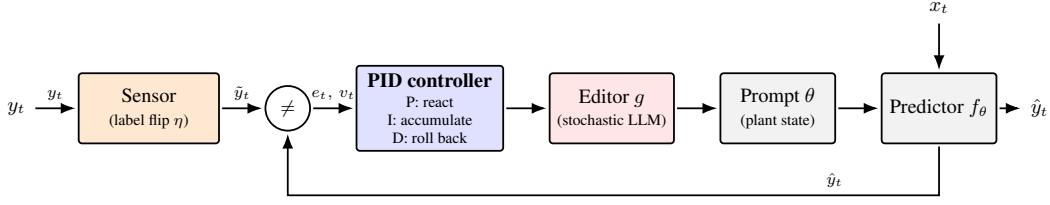
\begin{figure}[t]
\centering
\resizebox{\linewidth}{!}{%
\begin{tikzpicture}[
    node distance=6mm and 7mm,
    every node/.style={font=\small},
    plant/.style={draw, rounded corners=1.5pt, minimum height=11mm,
                  minimum width=18mm, align=center, thick, fill=gray!10},
    sensor/.style={draw, rounded corners=1.5pt, minimum height=11mm,
                   minimum width=22mm, align=center, thick, fill=orange!18},
    ctrl/.style={draw, rounded corners=1.5pt, minimum height=13mm,
                 minimum width=23mm, align=center, thick, fill=blue!12},
    act/.style={draw, rounded corners=1.5pt, minimum height=11mm,
                minimum width=19mm, align=center, thick, fill=red!10},
    sum/.style={draw, circle, minimum size=7mm, inner sep=0pt, thick,
                font=\footnotesize},
    arr/.style={->, >=Latex, thick, shorten >=1pt, shorten <=1pt}
]
\node (yt) {\small $y_t$};
\node[sensor, right=of yt]  (eval)  {Sensor\\ \scriptsize (label flip $\eta$)};
\node[sum,    right=of eval] (cmp)  {$\ne$};
\node[ctrl,   right=of cmp]  (ctrl) {\textbf{PID controller}\\[-1pt]
    \scriptsize P: react\\[-2pt]\scriptsize I: accumulate\\[-2pt]
    \scriptsize D: roll back};
\node[act,    right=of ctrl] (edit) {Editor $g$\\ \scriptsize (stochastic LLM)};
\node[plant,  right=of edit] (theta){Prompt $\theta$\\ \scriptsize (plant state)};
\node[plant,  right=of theta](pred) {Predictor $f_\theta$};
\node[right=4mm of pred]     (yhat) {\small $\hat{y}_t$};
\node[above=8mm of pred] (xt) {\small $x_t$};

\draw[arr] (yt)   -- node[above, midway] {\scriptsize $y_t$} (eval);
\draw[arr] (eval) -- node[above, midway] {\scriptsize $\tilde{y}_t$} (cmp);
\draw[arr] (cmp)  -- node[above, midway] {\scriptsize $e_t,\,v_t$} (ctrl);
\draw[arr] (ctrl) -- (edit);
\draw[arr] (edit) -- (theta);
\draw[arr] (theta)-- (pred);
\draw[arr] (pred) -- (yhat);
\draw[arr] (xt)   -- (pred.north);
\draw[arr] (pred.south) -- ++(0,-8mm) -| (cmp.south)
    node[pos=0.08, above] {\scriptsize $\hat{y}_t$};
\end{tikzpicture}%
}
\caption{The regulated textual feedback loop. Signals enter from outside: target $y_t$ from the left, input $x_t$ from above. The Sensor flips the label with probability $\eta$ (labels are either categorical, as in Parity, or bucketed categorical, as in Guess number / moving target). The comparator produces the observed error $e_t$ (binary for Parity, bucketed distance for numeric tasks); when the controller fires, editor $g$ rewrites the prompt $\theta$.}
\label{fig:system}
\end{figure}

Figure~\ref{fig:system} lays out the five-component closed loop. At each step $t$, a drifting environment presents input $x_t$ with hidden target $y_t$; the predictor emits $\hat{y}_t = f(\theta_t, x_t)$ under the current prompt $\theta_t$; a noisy sensor returns an observed label $\tilde{y}_t$ that disagrees with $y_t$ at rate $\eta$; the comparator then produces an observed error $e_t$ (binary $\mathbf{1}[\hat{y}_t \ne \tilde{y}_t]$ for categorical tasks, or a bucketed distance for numeric ones) and, on mismatch, a natural-language verbalization $v_t$. A PID controller reads the histories $\{e_\tau, v_\tau\}_{\tau\leq t}$ and decides whether to invoke the LLM editor $g$, which returns $\theta_{t+1} = g(\theta_t, \{v_\tau\})$. Because $g$ is a stochastic actuator with no gradient and $\theta$ is a persistent state whose bad edits compound, the controller's role is scheduling (\emph{when} and on what evidence to call $g$) rather than choosing the prompt itself; the goal is to track $y_t$ at low editing cost while remaining robust to corruption in $\tilde{y}_t$. In the guess-number game, for instance, $\theta_t$ encodes the agent's belief about the plausible value range as integer bounds $[L,U]$; on each turn ($x_t$), the LLM predictor $f$ emits a numeric guess ($\hat{y}_t$) conditioned on this belief; the environment scores it against a hidden target ($y_t$) and returns high/low/correct feedback with a distance bucket ($\tilde{y}_t$) that the sensor may flip; the comparator packages this into $(e_t, v_t)$, e.g., ``guess was too low, distance 10--19''; when the controller fires, the LLM editor $g$ rewrites $[L,U]$, tightening the belief for $\theta_{t+1}$.

\subsection{PID-inspired Gating}
\label{subsec:pid-gating}
We map each classical PID component onto a discrete editing gate. All three gates share the same eligibility check (unused call budget, error evidence since the last reset, no mandatory pause) and invoke the same LLM editor $g$, which receives the current prompt and up to $K$ recent observed examples (including correct ones; $K=8$ in our experiments); invalid proposals retain the prompt and every model call counts against the budget. The gate controls only \emph{when} $g$ is called, never the content of $\theta_{t+1}$.

\paragraph{Proportional Gate (P).} The proportional term reacts to the current error: the \emph{Proportional Gate} fires whenever the observer detects an error, i.e., $e_t=1$ in the binary case or $e_t\geq\tau_P$ when the error is a non-negative bucketed distance, and eligibility holds. It is the most responsive schedule while the most vulnerable to isolated corruption in $\tilde{y}_t$.
\begin{equation}
 a_t = \mathbf 1[e_t\geq\tau_P]\,\mathbf 1[\text{eligible at }t].
 \label{eq:reactive}
\end{equation}

\paragraph{Integral Gate (I).} The integral term accumulates evidence over time. The \emph{Integral Gate} maintains a decayed error sum and resets $A_t$ to zero after any attempted edit.
\begin{equation}
 A_t = \alpha A_{t-1}+e_t,\qquad
 a_t=\mathbf 1[A_t\geq\tau_I]\,\mathbf 1[\text{eligible at }t].
 \label{eq:gate}
\end{equation}

\paragraph{Derivative Gate (D).} The derivative term watches for post-edit deterioration. On firing, the \emph{Derivative Gate} snapshots the current prompt, holds the replacement fixed for $W$ observations, and restores the snapshot if the post-edit error mean rises by at least $\tau_D$ over the pre-edit mean. Full update order is in Appendix~\ref{app:algorithm}.
\begin{equation}
 r_t = \mathbf 1[\bar{e}_{\text{post}} - \bar{e}_{\text{pre}} \geq \tau_D].
 \label{eq:guarded}
\end{equation}

\section{Experiments}
\label{sec:protocol}
\subsection{Tasks}
\label{subsec:tasks}
We design three simple but representative tasks to test the self-evolving agent's capability across different output types (categorical and numeric) and drift patterns (sudden reversal, abrupt state shift, and smooth continuous drift). Together they probe how quickly the agent detects change, how it accumulates evidence under noisy feedback, and how gracefully it tracks a target that never stops moving. Hidden targets, boundaries, and initial rules are seed-randomized and never shown to the editor, so the study isolates LLM-mediated policy editing from end-to-end agent reasoning.
\begin{itemize}
\itemsep0em
\item \textbf{Parity} (categorical, sudden rule flips). Inputs are integers 1--10; the hidden rule assigns Red/Blue by parity, and the mapping reverses near steps 30, 60, and 90. Feedback is a categorical color, corrupted independently with probability $\eta$. Stresses categorical rule detection under discrete reversals.
\item \textbf{Guess number} (numeric, abrupt state shifts). The agent guesses a hidden integer in $[1,100]$ that shifts abruptly and periodically; feedback is high/low/correct paired with a distance bucket. Stresses numeric adaptation when the target state jumps.
\item \textbf{Track moving target} (numeric, smooth continuous drift). The hidden integer follows a smooth continuous trajectory (e.g., sinusoidal, rectangular, or exponential); our implementation uses a sinusoid completing $\sim\!1.5$ cycles over 120 steps, with no discrete change points. Stresses continuous tracking without visible boundaries a controller could anchor to.
\end{itemize}

\subsection{Experimental Setup}
\label{subsec:setup}
We evaluate seven policies (No Edit, Fixed Cadence, Proportional (P), Integral (I), Derivative (D), TextGrad, GEPA) on the three tasks under three feedback-corruption levels $\eta \in \{0, 0.1, 0.2\}$ with $T = 120$ steps per trajectory, using the same LLM as both predictor and editor; results are mean $\pm$ std over 5 seeds per cell. No Edit keeps the initial prompt unchanged for the full trajectory (zero edits, zero editor calls) and serves as a no-adaptation reference; Fixed Cadence fires an edit every 15 steps regardless of the error signal and serves as a data-independent control. Seeded inputs, hidden targets, and controller settings are paired across corruption levels for matched comparison. Evaluation metrics are task-specific: classification accuracy for \textbf{Parity} (higher is better), mean absolute error (MAE) for \textbf{Guess number} (lower is better), and root-mean-squared error (RMSE) for \textbf{Track moving target} (lower is better). Model calls are counted at every generation stage; tokens and deterministic candidate evaluations are reported separately. TextGrad and GEPA use the default parameters from their published libraries; interface and accounting details are given in Appendix~\ref{app:baselines}. We report primary results on \href{https://huggingface.co/Qwen/Qwen3-14B}{Qwen3-14B} (Table~\ref{tab:benchmark-main}) and a scaling-up ablation on \href{https://huggingface.co/Qwen/Qwen3.8-27B}{Qwen3.8-27B} (Table~\ref{tab:benchmark-27b}).

\section{Results and Analysis}
\label{sec:results}

\begin{figure}[H]
\centering
\includegraphics[width=\linewidth]{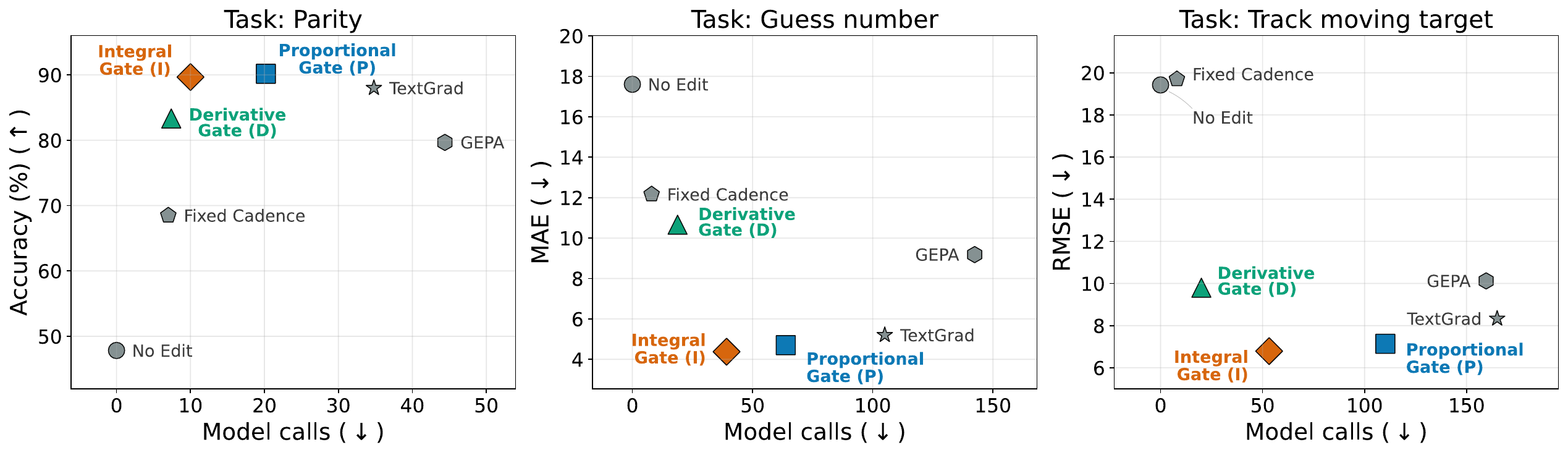}
\caption{Cost-quality snapshot at $\eta=0.1$ on Qwen3-14B (one marker per method, mean over 5 seeds). Higher accuracy and lower MAE/RMSE are better; fewer calls are cheaper. Points near the top-left (or bottom-left, for MAE/RMSE) indicate a better cost-quality trade-off.}
\label{fig:performance-vs-calls}
\end{figure}

\begin{table}[H]
\centering
\caption{Primary results on Qwen3-14B at $\eta=0.1$ ($10\%$ feedback corruption, 5 random seeds). Bold marks the best mean per column, excluding No Edit and Fixed Cadence.}
\label{tab:benchmark-main}
\setlength{\tabcolsep}{3pt}\scriptsize
\begin{tabular}{lrrrrrr}
\toprule
&\multicolumn{2}{c}{Parity}&\multicolumn{2}{c}{Guess number}&\multicolumn{2}{c}{Track moving target}\\
\cmidrule(lr){2-3}\cmidrule(lr){4-5}\cmidrule(lr){6-7}
Policy & Acc.(\%) $\uparrow$ & Calls $\downarrow$ & MAE $\downarrow$ & Calls $\downarrow$ & RMSE $\downarrow$ & Calls $\downarrow$\\
\midrule
No Edit & $47.8\pm5.8$ & $0.0\pm0.0$ & $17.61\pm3.05$ & $0.0\pm0.0$ & $19.42\pm2.05$ & $0.0\pm0.0$\\
Fixed Cadence & $68.5\pm10.1$ & $7.0\pm0.6$ & $12.17\pm3.54$ & $8.0\pm0.0$ & $19.70\pm4.10$ & $8.0\pm0.0$\\
\midrule
TextGrad & $88.0\pm4.4$ & $34.8\pm13.1$ & $5.20\pm1.47$ & $105.0\pm21.1$ & $8.33\pm1.57$ & $165.0\pm3.3$\\
GEPA & $79.7\pm6.4$ & $44.4\pm7.6$ & $9.18\pm1.55$ & $142.4\pm18.3$ & $10.12\pm0.91$ & $159.6\pm9.2$\\
\midrule
Proportional Gate (P) & $\mathbf{90.2\pm3.6}$ & $20.2\pm5.1$ & $4.70\pm2.76$ & $63.8\pm14.3$ & $7.14\pm1.42$ & $110.2\pm5.5$\\
Integral Gate (I) & $89.7\pm2.5$ & $10.0\pm3.0$ & $\mathbf{4.38\pm1.15}$ & $39.2\pm8.0$ & $\mathbf{6.79\pm2.38}$ & $53.2\pm1.2$\\
Derivative Gate (D) & $83.3\pm4.9$ & $\mathbf{7.4\pm2.1}$ & $10.66\pm2.19$ & $\mathbf{18.8\pm1.2}$ & $9.80\pm0.84$ & $\mathbf{20.0\pm0.0}$\\
\bottomrule
\end{tabular}
\end{table}

\begin{table}[H]
\centering
\caption{Scaling-up ablation on Qwen3.8-27B at $\eta=0.1$ ($10\%$ feedback corruption, 5 random seeds). Bold marks the best mean per column, excluding No Edit and Fixed Cadence.}
\label{tab:benchmark-27b}
\scriptsize\setlength{\tabcolsep}{3pt}
\begin{tabular}{lrrrrrr}
\toprule
&\multicolumn{2}{c}{Parity}&\multicolumn{2}{c}{Guess number}&\multicolumn{2}{c}{Track moving target}\\
\cmidrule(lr){2-3}\cmidrule(lr){4-5}\cmidrule(lr){6-7}
Policy & Acc.(\%) $\uparrow$ & Calls $\downarrow$ & MAE $\downarrow$ & Calls $\downarrow$ & RMSE $\downarrow$ & Calls $\downarrow$\\
\midrule
No Edit & $47.8\pm5.8$ & $0.0\pm0.0$ & $17.61\pm3.05$ & $0.0\pm0.0$ & $19.42\pm2.05$ & $0.0\pm0.0$\\
Fixed Cadence & $61.0\pm15.7$ & $7.2\pm0.7$ & $14.59\pm4.48$ & $7.8\pm0.4$ & $20.03\pm5.50$ & $8.0\pm0.0$\\
\midrule
TextGrad & $87.3\pm5.9$ & $34.8\pm13.9$ & $\mathbf{3.92\pm1.39}$ & $99.6\pm20.2$ & $7.03\pm1.22$ & $160.8\pm8.8$\\
GEPA & $79.7\pm6.4$ & $44.4\pm7.6$ & $8.27\pm1.75$ & $107.0\pm10.6$ & $10.31\pm0.98$ & $161.6\pm3.5$\\
\midrule
Proportional Gate (P) & $89.2\pm3.7$ & $22.6\pm6.4$ & $3.93\pm1.34$ & $67.6\pm19.5$ & $6.45\pm1.20$ & $108.0\pm3.9$\\
Integral Gate (I) & $\mathbf{89.7\pm3.7}$ & $10.0\pm3.7$ & $4.19\pm1.61$ & $39.8\pm10.1$ & $\mathbf{6.34\pm1.49}$ & $52.4\pm2.0$\\
Derivative Gate (D) & $82.3\pm4.3$ & $\mathbf{8.0\pm1.9}$ & $8.85\pm1.27$ & $\mathbf{19.2\pm1.0}$ & $10.25\pm2.14$ & $\mathbf{19.8\pm0.4}$\\
\bottomrule
\end{tabular}
\end{table}

Table~\ref{tab:benchmark-main} summarizes the primary comparison on Qwen3-14B at $\eta=0.1$: on every task, a PID gate attains the top quality metric among learned schedules ($90.2\%$ accuracy on Parity for Proportional; $4.38$ MAE on Guess number and $6.79$ RMSE on Track for Integral), while all PID gates issue fewer LLM calls than TextGrad and GEPA. The Integral Gate occupies a particularly efficient operating point: it matches or beats TextGrad on every quality metric while using roughly $2.7$--$3.5\times$ fewer editor calls, and the Derivative Gate uses the fewest calls on each task at a modest quality trade-off. The same cost-quality ordering replicates on a larger LLM (Table~\ref{tab:benchmark-27b}): on Qwen3.8-27B, the Integral Gate takes the top Parity accuracy ($89.7\%$) and the top Track RMSE ($6.34$) at roughly $3\times$ fewer calls than TextGrad, with Guess-number MAE within $0.3$ of the best. Absolute quality tracks editor capacity, but the Pareto ordering between PID gates and the optimizer baselines is preserved. Figure~\ref{fig:performance-vs-calls} visualizes the 14B snapshot: PID points sit up and to the left of the TextGrad/GEPA cluster, i.e., comparable or higher quality at substantially lower cost. This ordering is consistent across all three noise levels $\eta \in \{0, 0.1, 0.2\}$ (Appendix~\ref{app:noise}, Figure~\ref{fig:performance-vs-calls-all-noise}): the PID gates occupy a better cost--quality frontier than TextGrad and GEPA under both clean and corrupted feedback.

\section{Conclusion and Future Work}
\label{sec:conclusion}
We introduced PID-gated scheduling (Proportional, Integral, Derivative) that decides \emph{when} to invoke LLM prompt editors for continual prompt adaptation under drifting tasks and noisy feedback. Across three drifting tasks, the PID gates collectively yield a more favorable Pareto trade-off between task performance and editor-call budget than the baselines considered, typically matching the strongest optimizer at a fraction of its call cost (Section~\ref{sec:results}, Tables~\ref{tab:benchmark-main}--\ref{tab:benchmark-27b}, Figure~\ref{fig:performance-vs-calls}).
Future work includes the \emph{how-much} question (edit magnitude), for which a natural extension is an ``edit gain'' that limits the magnitude of the change from $\theta_t$ to $\theta_{t+1}$ so one controller regulates both frequency and magnitude; and online auto-tuning of the gate thresholds, borrowing established methods from classical control.
\label{body:end}
\clearpage
\bibliographystyle{plainnat}
\bibliography{references}
\clearpage
\appendix

\section{Update Order and Gate Characterization}
\label{app:algorithm}
\begin{algorithm}[h]
\caption{PID-inspired gate editing (Proportional, Integral, Derivative). One pass over $T$ steps; the gate $G$ selects the trigger condition and whether the post-edit regression check is armed. The Derivative gate reuses the Integral trigger ($A\geq\tau_I$) and adds a pre/post error-mean regression check that can roll back the edit.}
\begin{algorithmic}[1]
\Require prompt $\theta$; editor $g$; gate $G\in\{\mathrm{P},\mathrm{I},\mathrm{D}\}$; decay $\alpha$; thresholds $\tau_P,\tau_I,\tau_D$ ($\tau_I$ is shared by I and D); guard window $W$; history size $K$; call cap $B$; horizon $T$
\State $A\gets0$;\ \ $\mathcal H\gets\varnothing$;\ \ $\mathcal R\gets\varnothing$ (recent-error queue, length $W$);\ \ pending $\gets\varnothing$ (holds $(\theta_{\mathrm{snap}},\overline e_{\mathrm{pre}},\text{post-window})$ for D);\ \ cooldown$\gets0$;\ \ calls$\gets0$;\ \ evidence$\gets$\textsc{False} \Comment{True once any error has been seen since the last reset}
\For{$t=0,\ldots,T-1$}
\State Predict $\hat y_t=f(\theta,x_t)$; observe $\tilde y_t$; set $e_t=\mathbf{1}[\hat y_t\ne\tilde y_t]$ (categorical) or a bucketed distance (numeric)
\State Append $(x_t,\hat y_t,\tilde y_t,e_t)$ to $\mathcal H$; retain the latest $K$ entries
\State $\text{restored}\gets\textsc{False}$ \Comment{set if a D-gate rollback just happened; skips accumulator update}
\If{$G=\mathrm{D}$ \textbf{and} pending $\ne\varnothing$}\Comment{regression check}
   \State Append $e_t$ to pending's post-edit window
   \If{$W$ post-edit observations collected}
      \If{$\overline e_{\mathrm{post}}\geq\overline e_{\mathrm{pre}}+\tau_D$}
         \State Restore $\theta\gets\theta_{\mathrm{snap}}$; $A\gets0$; evidence$\gets$\textsc{False}; $\text{restored}\gets\textsc{True}$
      \EndIf
      \State pending$\gets\varnothing$
   \EndIf
\EndIf
\State Append $e_t$ to $\mathcal R$ (drop oldest if $|\mathcal R|>W$)
\If{\textbf{not} restored}
   \State $A\gets\alpha A+e_t$;\ \ \textbf{if} $e_t\geq 1$ \textbf{then} evidence$\gets$\textsc{True}
\EndIf
\If{cooldown $>0$} \State cooldown$\gets$cooldown$-1$; \textbf{continue} \EndIf
\State fire$\gets\bigl(G{=}\mathrm{P}\wedge e_t\geq\tau_P\bigr)\vee\bigl(G\in\{\mathrm{I},\mathrm{D}\}\wedge A\geq\tau_I\bigr)$
\If{fire \textbf{and} evidence \textbf{and} calls $<B$}
   \If{$G=\mathrm{D}$ \textbf{and} $|\mathcal R|=W$}
      \State pending$\gets\bigl(\theta,\ \overline e_{\mathrm{pre}}=\tfrac1W\sum \mathcal R,\ \text{empty post-window}\bigr)$
   \EndIf
   \State $\theta\gets g(\theta,\mathcal H)$;\ \ calls$\gets$calls$+1$
   \State $A\gets0$; evidence$\gets$\textsc{False};\ \ \textbf{if} $G=\mathrm{D}$ \textbf{then} cooldown$\gets W$
\EndIf
\EndFor
\end{algorithmic}
\end{algorithm}
Default settings (shared across gates): $\alpha=0.9$, $W=5$, $K=8$, $B=180$. Gate-specific: $\tau_I=1.5$ (used by I and D), $\tau_D=0.15$ (D only), $\tau_P=1$ (P only, binary-error case).

\clearpage
\section{Published Optimizers as Online Baselines}
\label{app:baselines}
TextGrad~\cite{yuksekgonul2025textgrad} and GEPA~\cite{agrawal2025gepa} use their pinned official packages, with inference routed exclusively through the common LLM-based editor. Both trigger after two observed errors since the previous invocation and require at least three remaining calls. Each restarts from the currently deployed prompt and receives at most eight past/current observations. Neither accesses clean evaluation metrics or future observations. These adaptations compare online use of the algorithms with measured resource costs, rather than reproducing their original task suites or claiming that all implementations are equivalent.

\paragraph{TextGrad.} Version 0.1.8 uses an explicit engine for each of three stages: \texttt{TextLoss} critiques the current structured prompt against the task specification and observed feedback; \texttt{backward} produces textual gradient feedback; \texttt{TextualGradientDescent} proposes an updated variable. Gradient memory is zero. Loss and backward stages request concise critiques of at most 100 words; the update retains the package's required variable tags. Deterministic prediction is outside the autograd graph: the optimized variable is the structured policy prompt itself. Missing tags or an invalid structured proposal retain the deployed prompt and consume all three calls.

\paragraph{GEPA.} Version 0.1.4 starts a fresh search population from the deployed prompt, uses instance-level Pareto parent selection, and permits at most three model reflections per invocation. Accepted proposals must strictly improve on a common training minibatch before full validation. Alternating history entries form disjoint training and validation subsets when at least four observations are available, with the newest in training; smaller histories reuse the same observations in both. Minibatches contain at most three training examples. Optional merging is disabled, and perfect-score minibatches may skip reflection. Validation is past noisy evidence, not an independent estimate of future task performance.

The common client counts all reflection calls. Candidate-example evaluations include rejected and malformed candidates and are logged separately. A nominal metric-evaluation limit provides an additional termination condition for skipped reflections; the official search checks it between iterations, so realized evaluations can exceed that nominal limit. Every proposal, selected parent, fitness evaluation, validation score, and final selection is retained. An invocation containing an invalid proposal may still deploy a different valid candidate; invalid-proposal count is therefore distinct from final prompt changes. Unexpected model or library failures halt the run rather than being relabeled successful no-ops.

\clearpage
\section{Robustness Across Feedback-Noise Levels}
\label{app:noise}

\begin{figure}[h]
\centering
\includegraphics[width=\linewidth]{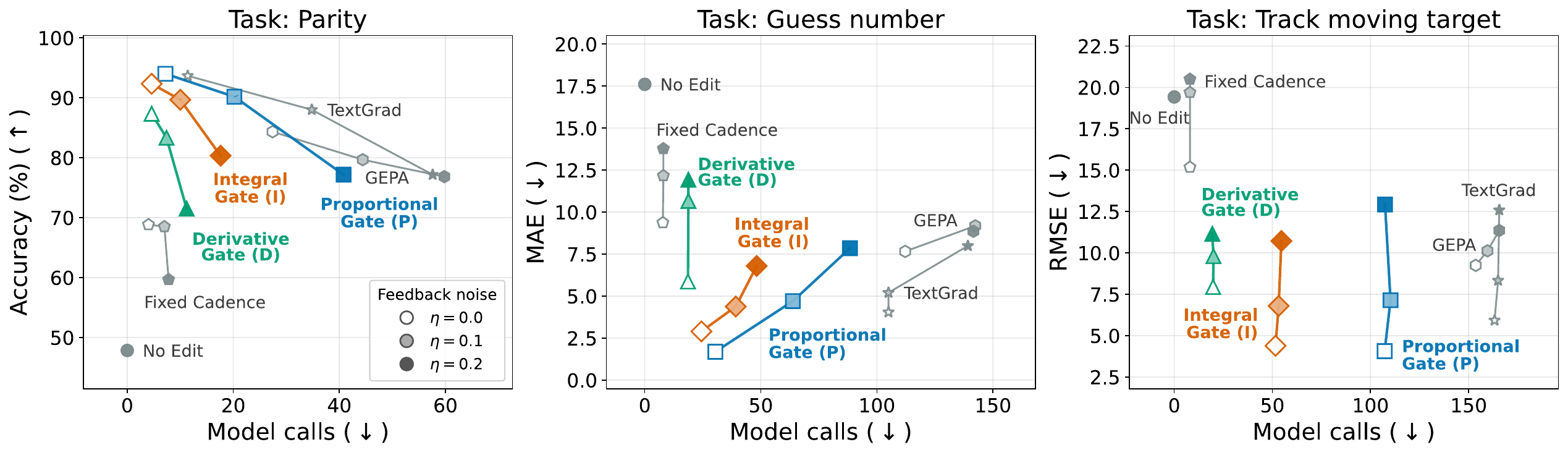}
\caption{Cost--quality tradeoff across all three feedback-noise levels on Qwen3-14B (mean over 5 seeds). Marker face shade encodes noise: hollow $\eta=0$, half-filled $\eta=0.1$, solid $\eta=0.2$. Lines connect the three noise points for each method. Points near the top-left (bottom-left for MAE/RMSE) indicate a better cost--quality trade-off. PID gates sit on a better frontier than TextGrad and GEPA at every noise level.}
\label{fig:performance-vs-calls-all-noise}
\end{figure}

Figure~\ref{fig:performance-vs-calls-all-noise} extends the primary comparison (Figure~\ref{fig:performance-vs-calls}) by plotting all three noise levels $\eta \in \{0, 0.1, 0.2\}$ together, with marker face shade encoding noise (hollow $\to$ half-filled $\to$ solid). Across every corruption level and on every task, the PID gates sit on a better cost--quality frontier than TextGrad and GEPA: they reach comparable or higher quality at a substantially lower editor-call cost. The Integral Gate, in particular, holds its operating point across the whole noise range, matching or beating the strongest baseline at a fraction of its call count. This replicates the Pareto conclusion of Section~\ref{sec:results} and shows that the ordering is not an artifact of the single noise level highlighted in the main text.

\clearpage
\section{Notation}
\label{app:notation}
Only recurring symbols are listed. One-shot symbols keep their meaning inline. The letter $g$ denotes the LLM editor throughout.

\begin{table}[h]
\centering
\caption{Recurring symbols.}
\label{tab:notation}
\small
\setlength{\tabcolsep}{6pt}
\renewcommand{\arraystretch}{1.15}
\begin{tabular}{lp{0.78\linewidth}}
\toprule
Symbol & Meaning \\
\midrule
$t,\,T$ & Step index and horizon; $T=120$ steps per trajectory. \\
$x_t,\,y_t$ & Input and hidden ground-truth target; $y_t$ never reaches the editor. \\
$\hat{y}_t,\,\tilde{y}_t$ & Predictor output $\hat{y}_t=f(\theta_t,x_t)$ and observed (possibly corrupted) feedback. \\
$e_t$ & Observed error: $\mathbf{1}[\hat{y}_t\neq\tilde{y}_t]$ for categorical tasks, or a bucketed distance for numeric ones. \\
$\theta_t$ & Current prompt (plant state); the editor's target of change. \\
$f,\,g$ & LLM predictor $\hat{y}_t=f(\theta_t,x_t)$ and LLM editor $\theta_{t+1}=g(\theta_t,\mathcal{H})$ (invoked when the gate fires). \\
$\mathcal{H},\,K$ & Rolling editor history: up to $K=8$ most recent observed examples. \\
$A_t,\,\alpha,\,\tau_I$ & Decayed error accumulator $A_t=\alpha A_{t-1}+e_t$; decay $\alpha=0.9$; threshold $\tau_I=1.5$. \\
$\tau_P,\,\tau_D$ & Proportional-gate threshold ($\tau_P=1$ in the binary case); Derivative-gate restoration threshold $\tau_D=0.15$. \\
$W$ & Guard window: number of observations used for the pre- and post-edit error means ($W=5$). \\
$B$ & Nonbinding call cap per trajectory ($B=180$). \\
$\eta$ & Feedback corruption probability (label-flip rate), $\eta\in\{0,0.1,0.2\}$. \\
\bottomrule
\end{tabular}
\end{table}

\end{document}